# Empirical Evaluation of RNN Architectures on Sentence Classification Task

Lei Shen, Junlin Zhang

Chanjet Information Technology

lorashen@126.com, zhangjlh@chanjet.com

**Abstract.** Recurrent Neural Networks have achieved state-of-the-art results for many problems in NLP and two most popular RNN architectures are "Tail Model" and "Pooling Model". In this paper, a hybrid architecture is proposed and we present the first empirical study using LSTMs to compare performance of the three RNN structures on sentence classification task. Experimental results show that the "Max Pooling Model" or "Hybrid Max Pooling Model" achieves the best performance on most datasets, while "Tail Model" does not outperform other models.

**Keywords:** RNN • LSTM • Sentence Classification

## 1      Introduction

Recurrent Neural Networks (RNNs), especially those Long Short-Term Memories (LSTMs) [9], are good at modeling varying length sequential data and have achieved state-of-the-art results for many problems in natural language processing, such as neural machine translation, question answering and text classification [2,3,12,14,19,20]. RNN is now the most popular method in sentence classification which is also a typical NLP task.

There are two widely used RNN structures in NLP tasks and we call them "Tail Model" and "Pooling Model" separately. However, there is no work focusing on comparing the performance of these various network architectures.

This paper presents the first empirical study using LSTMs to evaluate various RNN structures on sentence classification task. We also present a hybrid architecture that combines "Tail Model" with "Pooling Model" in this paper. Experimental results show that the "Max Pooling Model" or "Hybrid Max Pooling Model" achieves the best performance on most datasets, while "Mean Pooling Model" constantly has worse performance. And "Tail Model" does not outperform other models.

## 2    Related Work

A recurrent neural network [6] is introduced to process arbitrary length sequence and widely used in nowadays NLP tasks. Since simple recurrent network is difficult to train because the gradients will either vanish or explode [1], various improvements to the basic architecture were proposed and the Long Short Term Memory network is perhaps the most successful one. LSTM was originally introduced by Hochreiter and Schmidhuber [10] and in recent years some kinds of simplified LSTM were also introduced such as Gated Recurrent Units [3]. LSTM is used in our comparison since LSTM is comparable to GRU and the two both outperform basic RNN [4]. We use the LSTM structure that is introduced by Gers et al. [7] for comparison, since Greff et al. [9] evaluate the LSTM variants and find the model by Gers et al. performs comparable with other variants.

RNN has been successfully applied in NLP tasks and various RNN structures have been proposed. However, to our best knowledge, we didn't notice any work that evaluates the performance of the various RNN structures. So the following part of this paper will present the model comparison on sentence classification task.

## 3    Model

There exist two commonly used RNN structures in NLP tasks and we call them "Tail Model" and "Pooling Model" separately [15,19,20]. They both use RNN to provide feature layer for the fully connected layer in classification tasks. On the other hand, we regard the two different structures may be complementary for each other. So the third RNN structure is proposed in this paper and it will be called "Hybrid Model" in the following part of paper.

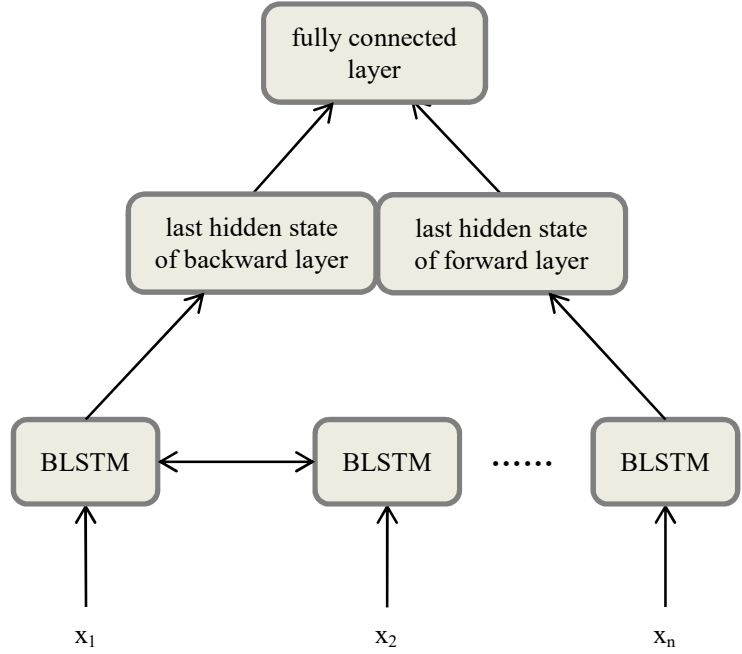

**Fig. 1.** Tail BLSTM Model

Figure 1 demonstrates the network structure of the "Tail BLSTM Model" in which the hidden layer is the concatenated hidden states of last node in the forward RNN and backward RNN. So the hidden layer is represented as following:

$$h = [\vec{h}, \overleftarrow{h}] \qquad (1)$$

Where $\vec{h}$ and $\overleftarrow{h}$ mean the hidden state of the last node in forward RNN and backward RNN respectively. It's obvious that $\vec{h}$ is the semantic mapping of sentence in forward direction and $\overleftarrow{h}$ contains the semantic information of sentence in reverse direction. $h$ can be regarded as the sentence feature and is fed into the next fully connected layer for classification:

$$y = f(Wh + b) \qquad (2)$$

Here $f$ is a softmax function in all of our experiments.

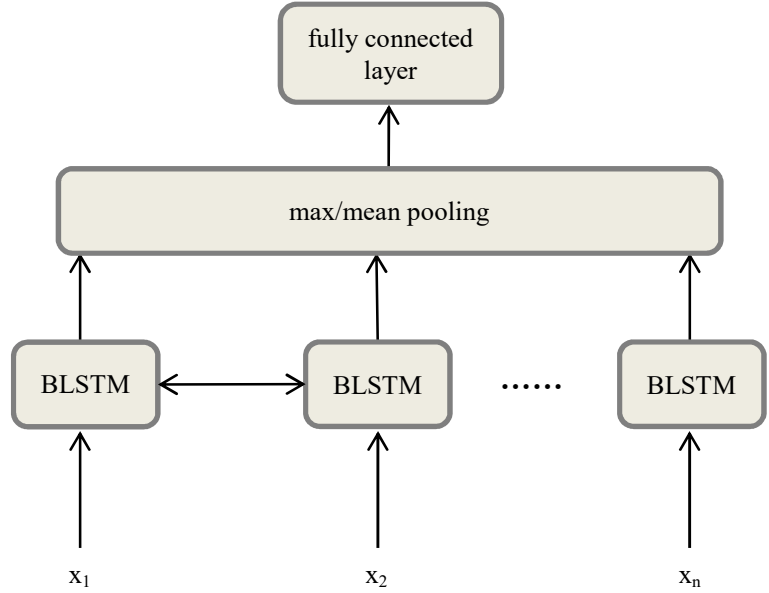

**Fig. 2.** Pooling BLSTM Model

The other widely used RNN structure is "Pooling Model" (Fig.2). The pooling BLSTM model can be regarded as each BLSTM node's voting for the feature layer and "mean pooling" is a common model for voting. "Mean pooling" calculates the value of the i'th position in vector $h$ by averaging the corresponding value of each hidden state vector from BLSTM nodes as following:

$$h_i^{\text{mean pooling}} = \frac{\sum_{j=1}^m h'_{ji}}{m} \quad (3)$$

Here we suppose the length of BLSTM sequence is m and $h'_j$ is the hidden state vector of the j'th BLSTM node.

The alternative for the "mean pooling" is "max pooling", which selects the max value of each position in all hidden state vectors. That implies the value of the i'th position in vector $h$ is calculated as:

$$h_i^{\text{max pooling}} = \max(h'_{ji}) \quad 1 \leq j \leq m \quad (4)$$

We didn't find any previous work that compares the performance of these two pooling methods(mean vs. max), so we also design some experiments to compare that.

The fully connected layer of "Pooling model" also can be represented by Eq. (2).

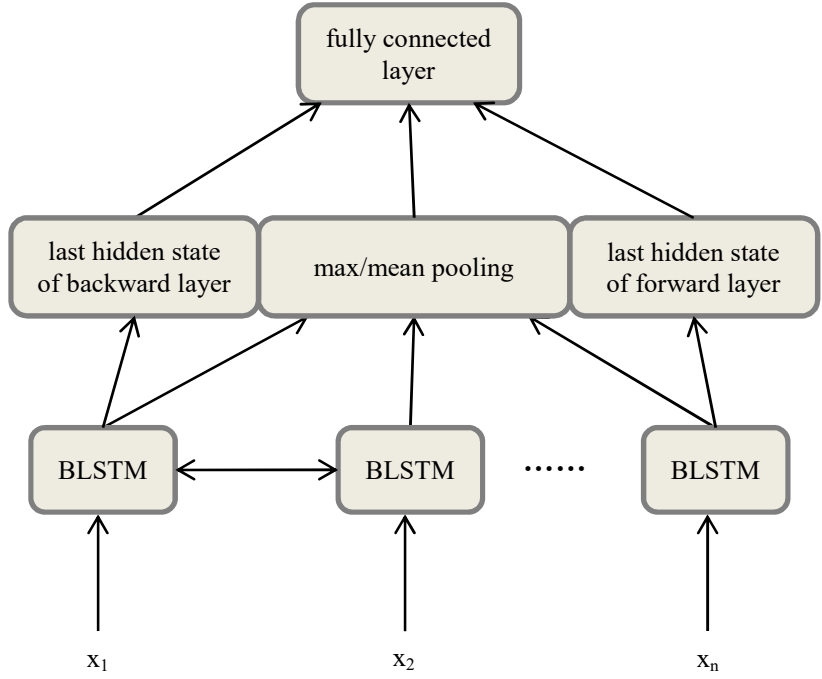

**Fig. 3.** Hybrid BLSTM Model

It's natural to speculate that the "Tail Model" and "Pooling Model" can be complementary for each other as there would be feature fusion. So we propose the third model called "Hybrid model" (Fig.3) which concatenates two different feature layers as following:

$$h = [\vec{h}, \overleftarrow{h}, h^{pooling}] \qquad (5)$$

The fully connected layer of "Hybrid Model" is also represented by Eq. (2).

The Tail, Pooling and Hybrid models based on LSTM are similar to BLSTM.

## 4  Datasets and Experimental Setup

### 4.1  Dataset

In order to evaluate the performance of the aforementioned three RNN structures, we design experiments on the following datasets for sentence classification task:

**MR**: The dataset of movie reviews with one sentence per review. The task is to detect positive/negative reviews [17].

**SST-1**: Stanford Sentiment Treebank—an extension of Movie Review but with fine-grained labels (very positive, positive, neutral, negative, very negative).The dataset is split into train/dev/test set and re-labeled by Socher et al. [18].

**SST-2**: Same data as SST-1 but only with binary labels. All neutral reviews are removed. The dataset is also split into train/dev/test set.

**Subj**: Subjectivity dataset. The task is to classify sentences into two categories: subjective and objective [16].

**TREC**: TREC question classification dataset, which classifies a question into 6 question types (person, location, numeric information, etc.). The dataset is split into train/dev/test set [13].

**CR**: Customer reviews of various products (cameras, MP3s etc.). The task is to predict positive/negative reviews [11].

The statistics of the datasets are in table 1.

Table 1. Statistics of datasets. CV means the dataset is not split into train/dev/test set, thus we use 10-fold cross validation.

| Dataset | Class | Dataset size | Test set size |
| --- | --- | --- | --- |
| MR | 2 | 10662 | CV |
| SST-1 | 5 | 11855 | 2210 |
| SST-2 | 2 | 9613 | 1821 |
| TREC | 6 | 5952 | 500 |
| Subj | 2 | 10000 | CV |
| CR | 2 | 3775 | CV |

### 4.2 Experiment Design

We design two groups of experiments: one is based on the BLSTM and the other is based on LSTM model. And the "Pooling Model" and the "Hybrid Model" each contains two experiments in order to compare the performance of max pooling and mean pooling strategy. Thus there are five experiments in each group for each dataset to evaluate the performance of these RNN architectures.

### 4.3 Parameters and Setup

We use word2vec vectors which were trained on 100 billion words from Google News as word embedding which can be accessed publicly. The vectors' dimensionality is 300 and those words not in the vectors are set randomly. We make the embedding static through all experiments to avoid the influence of the word embedding parameters in comparison. The initial weights of network parameters are drawn from a uniform distribution with standard deviation of 0.08 and the dropout is set to be 0.5.

Instead of achieving best performance for one single model, our focus is to compare the model's performance on sentence classification task under comparable condition. In order to make the parameters of the RNN models comparable, we set unidirectional LSTM's unit size of hidden layer to be 300 and bidirectional LSTM's unit size to be 185. Thus parameter size will be 7.2*100000 for both unidirectional and bidirectional LSTM in all experiments.

The Cross-Entropy criterion is used as loss function and stochastic gradient descent with Adagrad [5] is used for optimization. Gradients are computed through full BPTT for LSTM [8]. For those datasets that have no standard dev set, we randomly choose 10% of the training set as the dev set.

## 5   Results and Analysis

The experimental results are listed in the table 2(LSTM based models) and table 3(BLSTM based models).

Table 2. LSTM Experiment results on Datasets

| Model | MR | SST-1 | SST-2 | TREC | Subj | CR |
|---|---|---|---|---|---|---|
| LSTM_Tail | 79.0% | **46.3%** | 83.9% | 91.2% | 91.7% | 80.1% |
| LSTM_MeanPool | 78.1% | 44.4% | 83.4% | 88.2% | 91.1% | 79.0% |
| LSTM_MaxPool | 79.6% | 46.2% | **84.5%** | 91.8% | 91.6% | **82.4%** |
| LSTM_HybridMeanPool | 79.4% | 45.8% | 83.4% | 89.6% | **91.8%** | 81.0% |
| LSTM_HybridMaxPool | **79.9%** | 46.0% | 83.7% | **92.0%** | 91.7% | 82.3% |

Table 3. BLSTM Experiment results on Datasets

| Model | MR | SST-1 | SST-2 | TREC | Subj | CR |
|---|---|---|---|---|---|---|
| BLSTM_Tail | 78.7% | 45.0% | **84.1%** | **92.0%** | 91.8% | 80.4% |
| BLSTM_MeanPool | 79.9% | 45.8% | 83.5% | 90.2% | 91.8% | 81.2% |
| BLSTM_MaxPool | **80.4%** | **47.8%** | 83.7% | 91.8% | 92.1% | **83.0%** |
| BLSTM_HybridMeanPool | 79.2% | 44.5% | 83.4% | 91.0% | 92.0% | 80.2% |
| BLSTM_HybridMaxPool | 79.7% | 47.0% | 83.4% | 90.6% | **92.2%** | 82.5% |

First of All, we compare the performance of above-mentioned three kinds of models. One obvious conclusion is that the "Mean Pooling Model" constantly has worse performance than the others with large margin on most datasets, while "Max Pooling Model" or "Hybrid Max Pooling Model" achieves best performance on most datasets. The possible reason may be the max hidden state value which the max pooling method chooses contains the most meaningful information of the sentence. "Tail Model" does not outperform other models.

Obviously for the pooling strategy, max pooling method performs better than mean pooling method.

If we select LSTM_MeanPool model as a baseline LSTM benchmark to compare with the best run result, we can see the performance boost ranging from 0.44% to 4.37% on different datasets (Table 4). Similar performance improvements can also be observed in BLSTM models on all datasets (Table 5).

**Table 4.** Performance improvement of LSTM based Model (Best Run vs. LSTM_MeanPool)

| Model | MR | SST-1 | SST-2 | TREC | Subj | CR |
|---|---|---|---|---|---|---|
| LSTM_MeanPool | 78.1% | 44.4% | 83.4% | 88.2% | 91.1% | 79.0% |
| Best_Run | 79.9% | 46.3% | 84.5% | 92.0% | 91.8% | 82.4% |
| Performance Improvement | **+2.30%** | **+4.28%** | **+1.32%** | **+4.31%** | **+0.77%** | **+4.30%** |

**Table 5.** Performance improvement of BLSTM based model (Best Run vs. BLSTM_MeanPool)

| Model | MR | SST-1 | SST-2 | TREC | Subj | CR |
|---|---|---|---|---|---|---|
| BLSTM_MeanPool | 79.9% | 45.8% | 83.5% | 90.2% | 91.8% | 81.2% |
| Best_Run | 80.4% | 47.8% | 84.1% | 92.0% | 92.2% | 83.0% |
| Performance Improvement | **+0.63%** | **+4.37%** | **+0.72%** | **+2.00%** | **+0.44%** | **+2.22%** |

Out of our expectation, "Hybrid Pooling Model" does not constantly perform better than "Pooling Model" or "Tail Model". "Hybrid Pooling Model" only achieves best performance in 4 out of 12 results. This indicates that the two different RNN structures ("Tail model" and "Pooling model") didn't show the feature complementation function and the model fusion can't offer the performance boost.

The above experiment results analysis suggests we should give priority to the "Max Pooling Model" or "Hybrid Max Pooling Model" in sentence classification tasks given the above optional model choices. As "Tail Model" is widely used in NLP classification tasks, our experimental results provide another choice and the possibility to improve state-of-the-art result.

**Table 6.** BLSTM vs.LSTM model

| Model | MR | SST-1 | SST-2 | TREC | Subj | CR |
|---|---|---|---|---|---|---|
| Tail | L | L | B | B | B | B |
| MeanPool | B | B | B | B | B | B |
| MaxPool | B | B | L | B | B | B |
| HybridMeanPool | L | L | B | B | B | L |
| HybridMaxPool | L | B | L | L | B | B |

Secondly, we compare the performance difference of LSTM and BLSTM models. There are 30 different experiments (5 model*6 datasets) for both LSTM and BLSTM models. We list the winner of each experiment in Table 6 while "B" means BLSTM outperforms LSTM model and "L" means the vice versa. We can see from the comparison results that BLSTM outperforms LSTM model in majority of experiments. Though the performance boosts of several experiments are not large enough to show an overwhelming advantage, we believe that the BLSTM model has advantage because of the diversity of datasets and models in experiments. These experiment results probably indicate the BLSTM model can have better performance over LSTM model if both models use similar parameter settings and parameter number. We speculate this minor advantage may come from the BLSTM model's ability of capturing more features though these extra features don't have much effect on the model performance.

## 6      Conclusion

In this paper, we present the first empirical study using LSTMs to evaluate performance of various RNN structures on sentence classification task. We also present a hybrid architecture that combines "Tail Model" with "Pooling Model" in this paper. Experimental results show that the "Max Pooling Model" or "Hybrid Max Pooling Model" achieves the best performance on most datasets, while "Mean Pooling Model" constantly has worse performance than the others. And "Tail Model" does not outperform other models. As "Tail Model" is widely used in NLP classification tasks, our experimental results provide another choice and the possibility to improve state-of-the-art result.